\documentclass[conference]{IEEEtran}
\IEEEoverridecommandlockouts
\usepackage{cite}
\usepackage{amsmath,amssymb,amsfonts}
\usepackage{algorithmic}
\usepackage{graphicx}
\usepackage{textcomp}
\usepackage{xcolor}
\def\BibTeX{{\rm B\kern-.05em{\sc i\kern-.025em b}\kern-.08em
    T\kern-.1667em\lower.7ex\hbox{E}\kern-.125emX}}
\begin{document}

\title{Rethinking the semantic classification of indoor places by mobile robots*\\
\thanks{This work is funded by the Spanish Ministerio de Ciencia, Innovaci\'on y Universidades under projects HEROITEA (RTI2018-095599-B-C21) and ROBWELL (RTI2018-095599-A-C22); by the RoboCity2030-Madrid Robotics Digital Innovation Hub project (S2018/NMT-4331); and by the Wallenberg AI, Autonomous Systems and Software Program (WASP) funded by the Knut and Alice Wallenberg Foundation.}
}
\author{\IEEEauthorblockN{1\textsuperscript{st} Oscar M. Mozos}
\IEEEauthorblockA{\textit{AI for Life Lab, AASS} \\
\textit{\"Orebro University}\\
\"Orebro, Sweden \\
 oscar.mozos@oru.se}
\and
\IEEEauthorblockN{2\textsuperscript{nd} Alejandra C. Hernandez}
\IEEEauthorblockA{\textit{Robotics Lab, DISA} \\
\textit{Carlos~III University of Madrid}\\
Madrid, Spain \\
alejhern@ing.uc3m.es\\} 
\and
\IEEEauthorblockN{3\textsuperscript{rd} Clara Gomez}
\IEEEauthorblockA{\textit{Robotics Lab, DISA}  \\
\textit{Carlos~III University of Madrid}\\
Madrid, Spain \\
clgomezb@ing.uc3m.es}
\and
\IEEEauthorblockN{4\textsuperscript{th} Ramon Barber}
\IEEEauthorblockA{\textit{Robotics Lab, DISA} \\
\textit{Carlos~III University of Madrid}\\
Madrid, Spain \\
rbarber@ing.uc3m.es}
}

\maketitle

\begin{abstract}
 A significant challenge in service robots is the semantic understanding of their surrounding areas. Traditional approaches addressed this problem by segmenting the floor plan into regions corresponding to full rooms that are assigned labels consistent with human perception, e.g. \emph{office} or \emph{kitchen}. However, different areas inside the same room can be used in different ways: Could the table and the chair in my kitchen become my office
 ? What is the category of that area now? \emph{office} or \emph{kitchen}? 
 To adapt to these circumstances we propose a new paradigm where we intentionally relax the resulting labeling of semantic classifiers by allowing confusions inside rooms. Our hypothesis is that those confusions can be beneficial to a service robot. We present a proof of concept  in the task of searching for objects.
\end{abstract}

\begin{IEEEkeywords}
Semantic place categorization, Robotic Perception, Semantic Mapping, Object Search
\end{IEEEkeywords}

\section{Introduction}
Autonomous service robots are expected to perform a wide variety of everyday tasks in environments designed by and for humans. Therefore, they 
have to interpret the information about their surrounding environments in a similar way that people do. This is necessary to share perceptual and semantic concepts with users, to facilitate a natural communication with them and, in general, to improve their human-robot interaction capabilities~\cite{zender2008conceptual}. An important functionality in service robots working in indoor environments is the semantic categorization of places~\cite{kostavelis2015semantic, crespo2020mdpi}. In this problem, an autonomous robot should determine the semantic label of the area where it is located. This label should be consistent with the human perception about the same area and, therefore, it should share the same semantic meaning  with the user. Example labels for indoor areas include \emph{office} or \emph{kitchen}. 

\begin{figure}[t]
	\centering
	\includegraphics[width=\columnwidth]{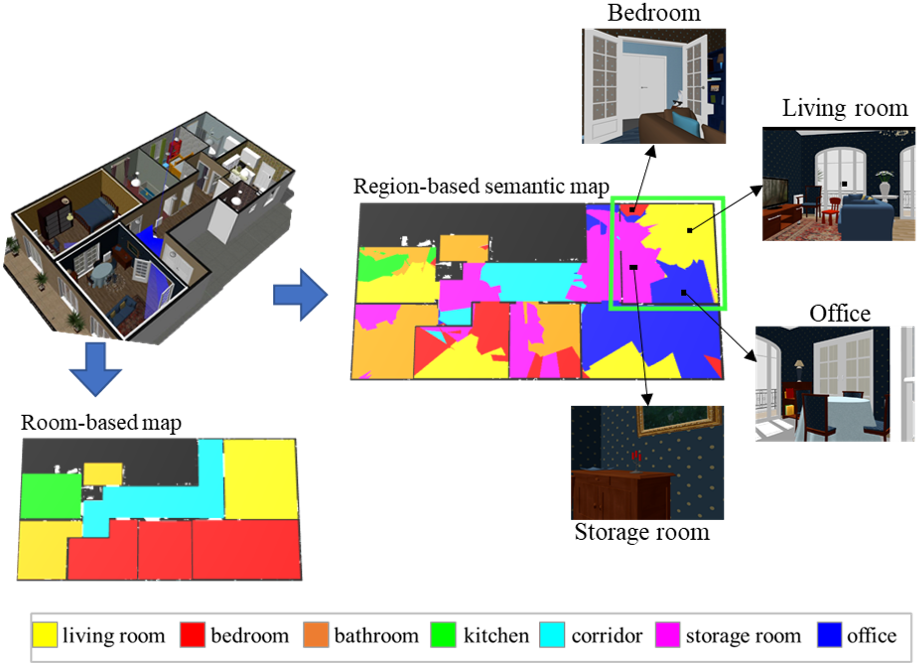}
	\caption{Upper-left: the original indoor environment.  Bottom-left: the semantic division based on full rooms. Right: example of our semantic division that keeps confusions based on appearance.}
	\label{fig:motivation}
\end{figure}

Traditional methods for constructing 2D semantic maps of indoor environments aim to assign a semantic label to each full room of a given environment, and  
efforts are directed to get a clean room-based map like the one shown in Fig.\ref{fig:motivation}  \cite{kostavelis2015semantic, crespo2020mdpi,qi2020building,sunderhauf2016icra,goeddel2016learning}.
However, the recent pandemic events have made some people change the way of perceiving or using their environments, mainly due to the shift to online activities from home. As a consequence, people have changed the meaning of some spaces inside their homes by assigning different uses to the same area. For example, the table and chairs in the kitchen can be seen or used as an office. 


Inspired by those circumstances, we present a new paradigm for the problem of semantic place categorization by mobile robots. The main idea 
is to intentionally keep confusions in the classification of subareas inside the rooms that result from the perception system of the robot. 
Our claim is that those confusions can naturally correspond to real alternatives, like for example the table and the chair inside the kitchen that can be classified as a small office inside that kitchen. Moreover, those confusions can improve particular tasks carried out by a service robot. 


This paper presents a summary of our  first approximation to transfer this change of paradigm into the perception system of a mobile robot working in an indoor environment \cite{alejandra2023ras}. Our approach creates a subdivision of the environment by maintaining the confusions 
of the perception system of the robot
which are due to the scene appearance or to the distribution of objects.
We do not aim to create new classifiers for this new categorization task. Instead, we use standard classifiers and rely on their confusions to obtain our subdivisions. An example is presented in the right part of Fig.~\ref{fig:motivation} in which a full room is divided into different subregions that correspond to the output of the appearance classifier from different viewpoints. 

As a proof of concept we implemented a robotic task consisting of searching daily life objects in simulated and real-world scenarios 
that
showed that keeping smaller confusing regions increased the efficiency in the search.

\section{Methodology and experimental results}
To create our maps we use the output of two different classifiers. The appearance-based classifier is based on a deep VGG16 architecture trained on the Places365 database \cite{zhou2017places}. The object-based map is based on histograms of objects that are found in different indoor spaces from the NYU depth V2 dataset \cite{silberman2012indoor}. Each classifier output results in a different semantic map, appearance-based and object-based, which are merged in a later stage. For specific details on the implementation we refer the reader to our original work \cite{alejandra2023ras}.

\begin{figure}[t]
 	\centering
 	\includegraphics[width=\columnwidth]{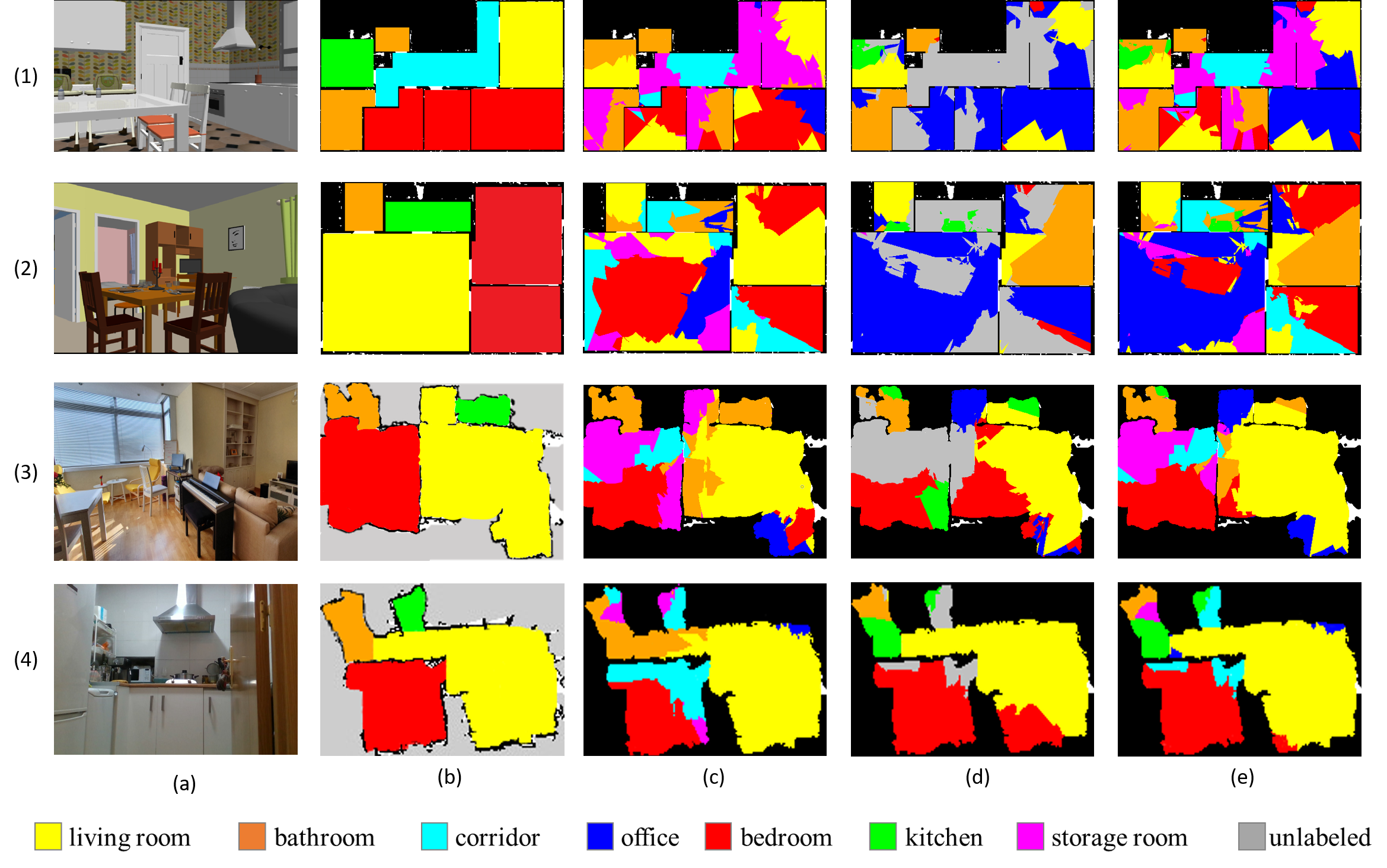}
  	\caption{Rows (1) and (2) show simulated environments while rows (3) and (4) depict real ones. Columns: (a) example scene image; (b) full 2D map; (c) appearance-based confusion map; (d) object-based confusion map; (e) merged confusion map.}
 	\label{fig:resulting_maps}
 \end{figure}

Fig.\ref{fig:resulting_maps} shows the results from our semantic segmentation that includes confusions based on appearance and on the distribution of objects applied to two simulated and two real home environments.    
We can see the difference between a manually segmented full room division (b) and the subdivisions obtained keeping the confusing areas (c-d-e). Many of those confused subdivisions make sense and correspond to potential real alternatives. 
For instance, in Fig.\ref{fig:resulting_maps} (1e) we can see how the living room (upper right room) is divided into several subregions: an office (blue), a storage room (magenta) and a living room (yellow), which make sense according to the contextual information and objects present in each subarea.

In a next experiment, we compare the performance of our robot in a object searching task. 
We compare our results with the work in \cite{hernandezefficient} (our baseline), where full rooms are assigned a single place category.

\begin{figure}[thb]
	\centering
	\includegraphics[width=\columnwidth]{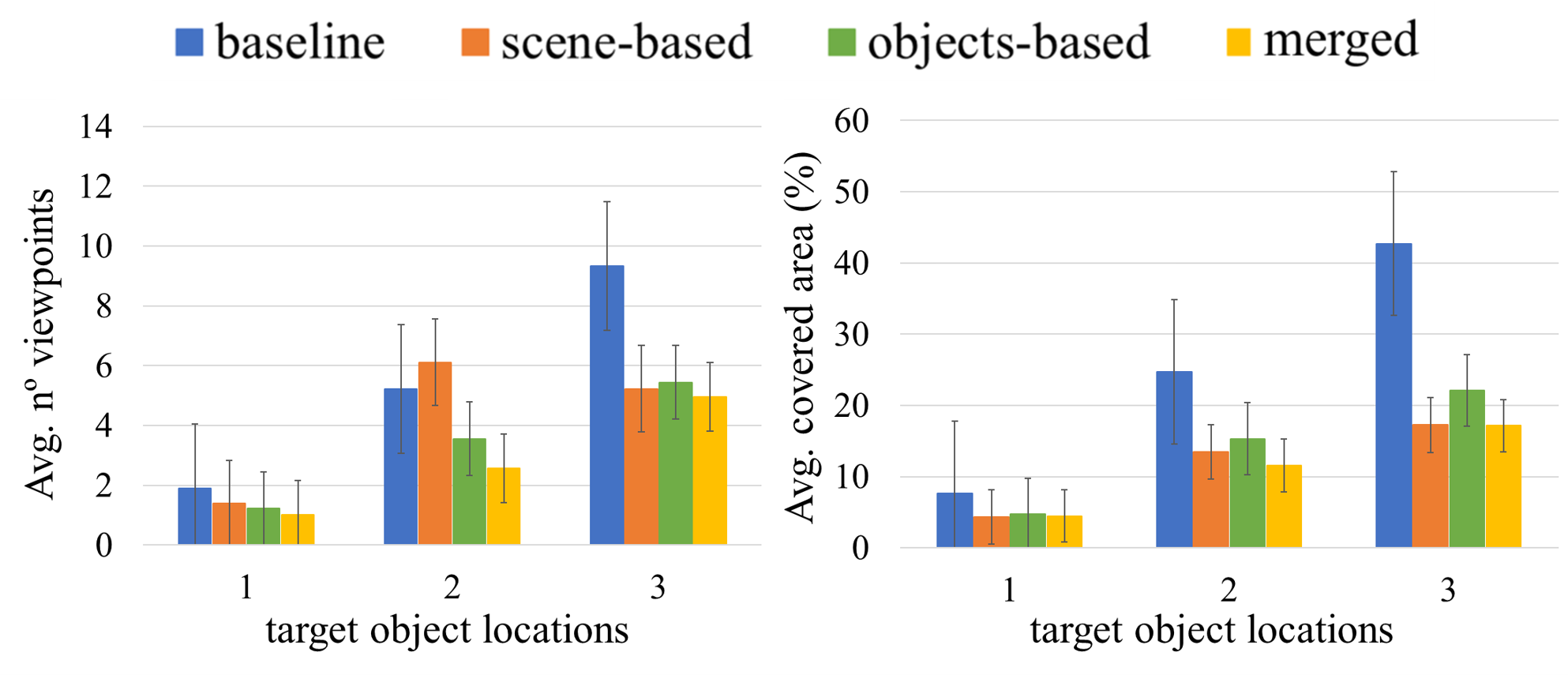}\\
        \vspace{-2mm}
        (a) Simulated environment \\
        \vspace{5mm}
	\includegraphics[width=\columnwidth]{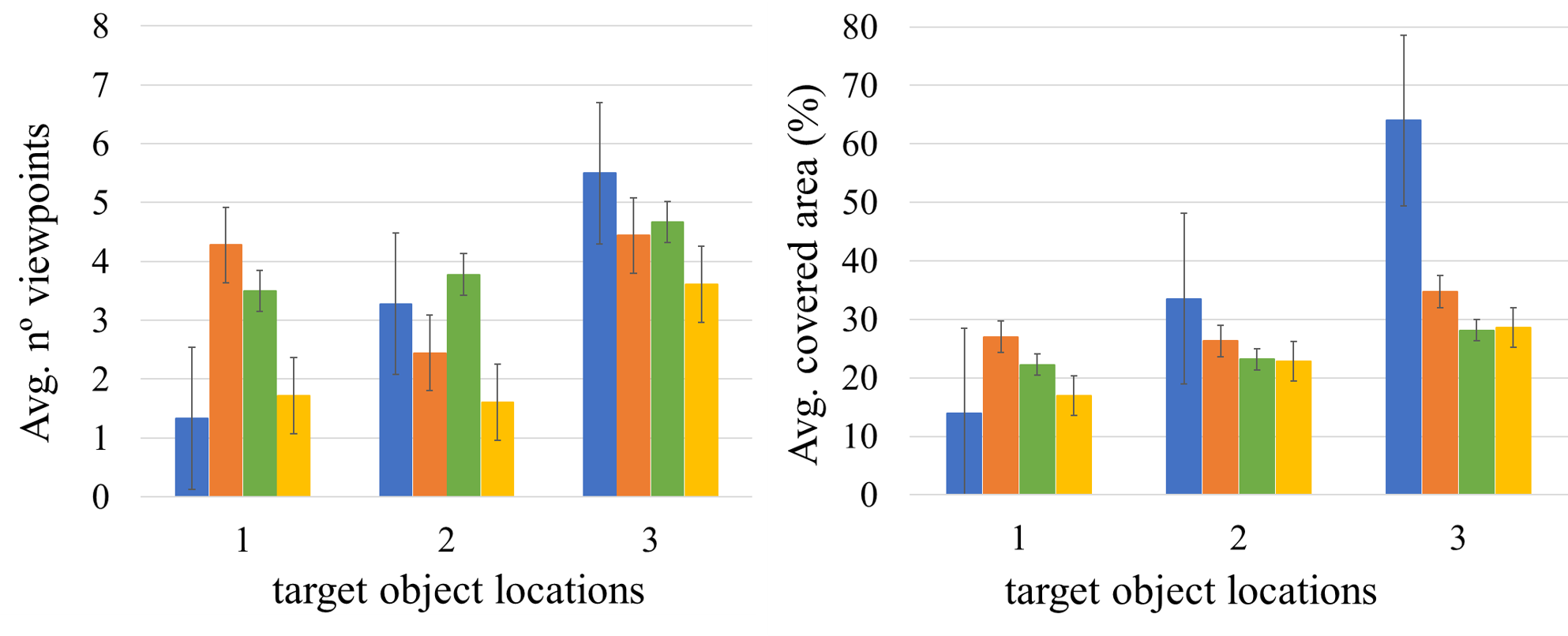}\\
        \vspace{-2mm}
        (b) Real-world environment \\	
	\caption{Results considering different initial locations for the lost object according to prior probabilities. X-axis values: 1=most probable place; 2=second most probable place, 3=third most probable place. Our approach (yellow) improves the search task by keeping the covered area  and the viewpoints visited at a low level both in simulated and real-world scenarios.}%
	\label{fig:graph2_locations}%
\end{figure}

We located the searched objects in three different places in decreasing order of their prior probability of being found, i.e., the first place has the highest probability of containing the object, the second place has the second highest probability, and so on. Location probabilities were obtained from statistics on object-scene co-occurrences of daily life objects \cite{hernandezefficient,brucker2018semantic}. For each object, we repeated the search 3 times in each different probable place. Results are shown in Fig.\ref{fig:graph2_locations}, where we show the number of viewpoints and the covered area needed by the robot to find the object. As the object is located in a less probable place, those values increase in the baseline strategy that uses full room maps~\cite{hernandezefficient}, while our approach based on confusions keeps lower numbers. This is an important result since, usually, lost objects are not found in the most probable place, and that is why it is more difficult to find them. Our approach based on confusion maps provides better efficiency in those cases. Further experiments are shown in \cite{alejandra2023ras}.

\section{Conclusion}
We presented a change of paradigm in the problem of semantic place categorization in indoor environments. Our claim is that confusions in the perception system of the robot can naturally correspond to real alternatives, i.e a small office inside a kitchen, and thus they can improve particular robotic tasks. Further improvements aim to study compatible confusions, e.g. a kitchen can be an office, versus incompatible ones, e.g. a corridor cannot be a toilet; and the integration of spatial-temporal human activities in different areas. 

\bibliographystyle{elsarticle-num}
\bibliography{mybibfile}

\end{document}